\documentclass[11pt]{article}

\usepackage[final]{acl}
 
\usepackage{times}
\usepackage{latexsym}
\usepackage{booktabs}
\usepackage{multirow}
\usepackage{tabularx}
\usepackage{float}
\usepackage[T1]{fontenc}

\usepackage[utf8]{inputenc}

\usepackage{microtype}
\usepackage{comment}
\usepackage{inconsolata}

\usepackage{graphicx}
\usepackage{amsmath}
\title{Reading Between the Layers: Cross-Modal Alignment and Representational Hierarchy in Vision Transformers}

\author{
    Sugam Karki$^{1}$ \qquad
    Raúl Vázquez$^{2}$ \qquad
    Yova Kementchedjhieva$^{1}$ \\
    $^{1}$MBZUAI \qquad
    $^{2}$University of Helsinki \\
    \small\textbf{Correspondence:} \texttt{sugam.karki@mbzuai.ac.ae} \quad  \texttt{yova.kementchedjhieva@mbzuai.ac.ae}
}

\begin{document}
\maketitle
\begin{abstract}
While large vision-language models excel at optical character recognition, it remains unclear how their underlying vision encoders process text and its meaning, and whether they support anything resembling reading comprehension. We systematically study transformer-based vision encoders, using cross-modal alignment between rendered text and visual concepts together with probing and retrieval-based analyses as a controlled window into these broader capabilities. Our layer-wise analyses span orthographic, grammatical, and semantic organization, and reveal that strong cross-modal mapping emerges only in the deepest layers, after unimodal object and text recognition have already developed, coinciding with a progression from surface-level to semantic representations. Although this hierarchy is strongest in contrastively trained models, larger self-supervised models and text-rich training data partially support similar behaviors without explicit language supervision. Together, our results suggest that increasing model and training-data scale, as well as text-rich training data, can partially compensate for the absence of language supervision.

\end{abstract}

\section{Introduction}

In human literacy research, a distinction is often drawn between \textit{decoding}, the ability to recognize written symbols, and \textit{reading comprehension}, the ability to connect those symbols to semantic meaning \cite{gough1986decoding, hoover1990simple}. Much like novice learners of English who recognize ``horse'' across contexts without grounding it in the concept of the animal itself, modern vision encoders exhibit decoding (a.k.a.~OCR) capabilities in their processing of in-image text \citep{lin2024parrotcaptionsteachclip}. While some studies suggest that transformer-based vision encoders (ViT) develop representations supporting cross-modal semantic alignment between text and visual concepts \citep{wang2025textural,materzynska2022disentanglingvisualwrittenconcepts}, the scope of these capabilities and their underlying representational dynamics have not been systematically studied. In this work, we use cross-modal alignment between rendered text and visual concepts, together with probing and retrieval-based analyses, as a controlled window into these broader reading comprehension capabilities.

\begin{figure}[t]  
\centering      
\includegraphics[width=\linewidth, trim=0.8cm 0.45cm 0.25cm 0.25cm, clip]{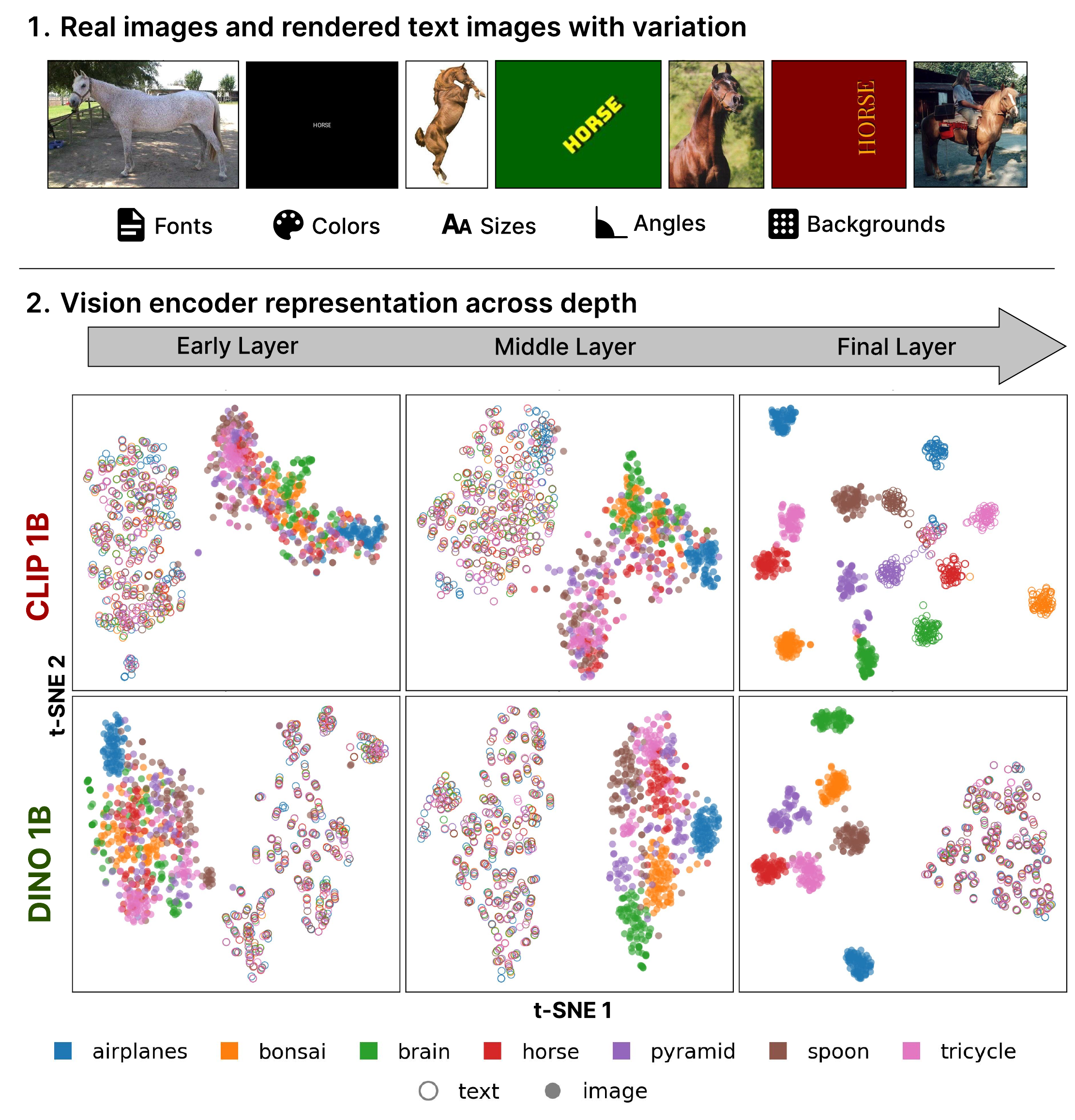}
\caption{\textit{Top:} Real images and rendered text variants used for cross-modal probing. \textit{Bottom:} t-SNE projections of CLIP (1B) and DINO (1B) representations across depth. In CLIP, same-concept representations from both modalities progressively align, while DINO maintains modality separation throughout.}
\label{fig:summaryfigure}  
\end{figure}  

Motivated by the distinction between surface decoding and broader reading capabilities, we study whether cross-modal semantic alignment builds upon visual object recognition or develops independently across network depth. As shown in Figure \ref{fig:summaryfigure}, we extend standard probing methodologies \citep{rogers2020primerbertologyknowbert, conneau2018sentevalevaluationtoolkituniversal} to ViT encoders by rendering textual stimuli as images with diverse visual augmentations. Using these stimuli, we conduct layer-wise probing to examine whether linguistic representations in vision encoders are organized hierarchically across layers, progressing from surface form to grammatical regularities and finally semantic understanding, analogous to hierarchies observed in humans \cite{friederici2002towards} and language models \cite{jawahar2019does, tenney2019bert}. To support these probe-based findings, we also evaluate cross-modal alignment and retrieval between rendered text and images, allowing us to compare when shared cross-modal structure emerges relative to unimodal object and text recognition. Finally, we address the role of training signals: while language supervision and data composition are known to shape surface-level reading \citep{lin2024parrotcaptionsteachclip, fan2025scalinglanguagefreevisualrepresentation, liu2025datalanguagesupervisionmakes}, we study if they similarly govern the development of deeper grammatical and semantic organization. 

Through a systematic study via layer-wise probing, supported by controlled cross-modal alignment experiments, spanning surface-level character recognition, grammatical regularities, and semantic alignment, we address these questions across three training paradigms — contrastive (CLIP) \cite{radford2021learningtransferablevisualmodels}, self-supervised (DINO) \cite{caron2021emergingpropertiesselfsupervisedvision}, and masked autoencoding (MAE) \cite{he2021maskedautoencodersscalablevision} — to understand how training objective modulates each:

\noindent\textbf{RQ1.} How and when does cross-modal semantic alignment emerge in vision encoders?

\noindent\textbf{RQ2.} What representational structure underlies this alignment?

\noindent\textbf{RQ3.} How do scale and data composition shape its development?

Our results reveal three main findings. First, strong cross-modal semantic alignment between rendered text and visual concepts emerges only in the deepest layers, after unimodal object and text recognition have already developed. Second, this alignment coincides with a representational hierarchy, clearest in CLIP, in which orthographic information emerges early, grammatical regularities peak at intermediate depths, and semantic organization consolidates in the deepest layers. Third, in self-supervised encoders, increasing model capacity from 300M to 2B mainly strengthens surface-level representations, whereas the 7B model, which is also trained on more data, develops a clearer hierarchy and measurable cross-modal transfer; enriching training data with text produces more modest improvements.

\section{Background}

\subsection{From Decoding to Cross-Modal Semantic Alignment in Vision Encoders}

Vision-language models trained using contrastive learning can align natural-language descriptions with visual concepts at the level of whole images \citep{radford2021learningtransferablevisualmodels}. Yet compositional image-text evaluations reveal limited sensitivity to word order, relations, and other fine-grained linguistic distinctions \citep{thrush2022winoground, yuksekgonul2022and}. Because these studies assess the final cross-modal embedding space, they do not isolate how an image encoder processes written language presented as pixels.

Whether ViTs go beyond decoding to develop cross-modal semantic alignment between written words and visual concepts has received less attention than surface-level character recognition. \citet{materzynska2022disentanglingvisualwrittenconcepts} provide evidence that they can, showing that CLIP's image encoder matches rendered images of words to natural images of the concepts those words describe. \citet{wang2025textural} extend this picture with a layer-wise analysis, finding that the early layers process text as visual texture while semantic differentiation appears only in the final layers, though they argue that improvements are largely driven by orthographic disambiguation rather than abstract meaning, casting doubt on whether these representations reflect semantic grounding beyond orthographic disambiguation, even in the deepest layers.

Consistent with this, \citet{lemesle2022languagebiasedimageclassificationevaluation} find that semantic word representations in CLIP's visual processing are not shared with image representations, with word form dominating over visual meaning for word-embedded images. \citet{park2025emergencetextreadabilityvision} further show that surface text reading ability develops abruptly during training while semantic alignment develops more gradually, suggesting that the two capabilities are not only functionally distinct but follow different developmental trajectories. To the best of the authors' knowledge, prior studies have left questions about the emergence of language-relevant representations in image encoders unanswered. Specifically, it remains unclear whether image encoders develop these representations, whether the representations underlying cross-modal semantic alignment follow a principled surface-to-semantic hierarchy, and how these representations relate to the encoder’s broader visual capabilities. We address these gaps through controlled layer-wise probing and cross-modal alignment experiments.

\subsection{Hierarchical Organization Across Layers}

Prior work on probing pretrained language models reveals a consistent hierarchical organization across layers, from surface features to grammatical structure to semantics \citep{rogers2020primerbertologyknowbert, jawahar2019does, tenney2019bert}. \citet{tatariya-etal-2024-pixology} extend this to PIXEL \citep{rust2023language}, a ViT-based language model trained on rendered text, finding a similar surface-to-semantic progression, although notably absent in ViT-MAE, which lacks language supervision. \citet{dumpala2024seeing} further show that VLM text encoders exhibit weaker syntactic organization than unimodal language models, with pre-training objective as the primary driver. Our work extends this line of inquiry to vision encoders processing rendered text through their image encoder, asking whether hierarchically organized language-relevant structure arises in models never trained on language.

\subsection{Role of Training Paradigm, Data, and Scale}

Learned model behavior is shaped not only by architecture and training objective, but also by the composition and scale of pretraining data. Visual datasets can contain incidental correlations that models exploit as shortcuts \citep{torralba2011unbiased, geirhos2020shortcut}; in self-supervised learning, data curation likewise affects the invariances and semantic distinctions captured by learned representations \citep{purushwalkam2020demystifying, van2021revisiting}. These findings motivate studying which statistical regularities support visual text processing.

This question is particularly salient for reading-related behavior in vision encoders. \citet{lin2024parrotcaptionsteachclip} show that CLIP's text recognition arises largely from a shortcut: roughly half of training images contain embedded text, and the majority of these have captions that repeat the same text, leading the model to associate visual character patterns with their transcriptions rather than principled reading. \citet{liu2025datalanguagesupervisionmakes} disentangle the contributions of data composition and language supervision more directly, finding that text-rich training data accounts for a substantial portion of the OCR gap between CLIP and DINO, with language supervision providing an additional benefit. Building on this, \citet{fan2025scalinglanguagefreevisualrepresentation} show that the remaining advantage of language supervision on surface-level reading can be further offset by increasing model scale. Whether these factors---supervision, data composition, and scale---similarly govern the emergence of cross-modal semantic alignment, as measured through cross-modal alignment between written words and their visual referents, remains an open question and the primary motivation for RQ3.

\begin{figure}[t]  
\centering      
\includegraphics[width=\linewidth]{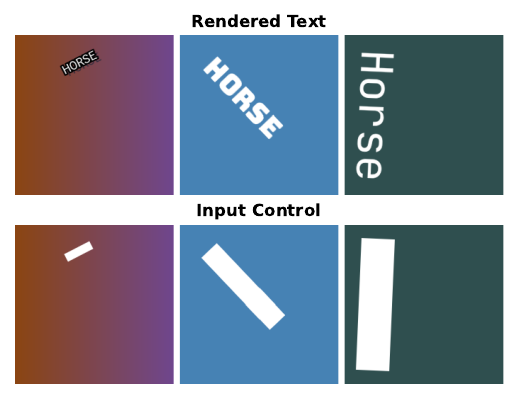}
\caption{Rectangle input controls for three stylistic text renders of the word ``horse''. The control follows the same rendering style as the text.}
\label{fig:input_control}                             
\end{figure}

\section{Methodology}

A central challenge in studying visual text processing in vision encoders is isolating textual representations from broader visual context, since vision encoders operate directly on pixels and do not accept discrete text tokens as input. Moreover, real images containing text introduce uncontrolled variation in layout, context, and background, making it difficult to isolate the mechanisms underlying visual text processing.
We adopt a controlled evaluation strategy in which textual inputs are rendered as images (Figure~\ref{fig:input_control}, top row). This allows us to present identical linguistic content in diverse visual forms by changing typographic features such as font, size, and layout. These variations reduce reliance on fixed low-level visual cues, allowing us to assess whether representations capture more generalizable textual structure. An additional advantage of this strategy is that rendered text images naturally serve as a \textit{cross-modal bridge} between visual and textual representations. We leverage this in our cross-modal experiments by pairing natural images with rendered text images of their corresponding class labels, allowing us to compare how shared concepts are represented across modalities while controlling for visual form.

\paragraph{Representation extraction.}
\label{sec:rep-ext}
We analyze frozen representations across network depth, extracting the post-LayerNorm residual stream from every transformer block. At each layer, we consider both the CLS token and mean-pooled patch tokens (AVG). For MAE models, we set the mask ratio to zero at inference so that all patch tokens are present. These representations form the basis of the task-specific analyses introduced below, where we define the inputs, prediction targets, and datasets for each research question.

\paragraph{Models.}
We study vision encoders representing three distinct pre-training paradigms: contrastive learning (CLIP; \citealt{radford2021learningtransferablevisualmodels}), self-supervised learning (DINO; \citealt{oquab2023dinov2}), and masked autoencoding (ViT-MAE; \citealt{he2021maskedautoencodersscalablevision}). To ensure a fair comparison, we follow \citet{fan2025scalinglanguagefreevisualrepresentation} and use models of comparable scale ($\approx$1B parameters) trained on the same data.\footnote{{\scriptsize With HuggingFace handles: \url{facebook/webssl-dino1b-full2b-224,} \url{/webssl-mae1b-full2b-224, /metaclip-h14-fullcc2.5b}}} We additionally examine scaling effects using 300M, 2B, and 7B DINO models. All models use the ViT architecture. During probe training, the model weights are kept frozen and the weights of the respective classifiers are adjusted. Freezing the weights helps to ensure that the probing measures properties of the pretrained representations without altering them  via task adaptation \citep{hewitt2019designinginterpretingprobescontrol}.  Appendix~\ref{appx:models} lists full model details.

\section{RQ1: Cross-Modal Semantic Alignment}
\label{sec:rq1}

\subsection{Tasks}
\label{subsec:tasks}
\paragraph{Cross-modal classification.}

Cross-modal classification tests whether object categories can be decoded across modalities from a shared representation space. The input is a frozen layer representation of either a natural image or a rendered class label, and the prediction target is the corresponding object category. We pair natural images from Caltech-256 \citep{griffin2007caltech} (143 classes, 50 images per class) and ImageNet-1K \citep{deng2009imagenet} (1,000 classes, 50 validation images per class) with rendered text images of their class labels, generated across diverse fonts, sizes, and backgrounds. At each layer, we train a logistic regression classifier on representations from one modality and evaluate it on representations from the other, averaging accuracy across the two transfer directions. We also evaluate within-modality classification to compare cross-modal transfer with unimodal object and text recognition.

Because probing tasks vary substantially in label cardinality, raw accuracy is not directly comparable across tasks, layers, and models. We normalize using \textbf{Percentage Accuracy Gain (PAG)} \citep[e.g.][]{Goodfellow-et-al-2016}:
\[
\text{PAG} = \frac{\text{Acc} - \text{Acc}_{\text{base}}}{100 - \text{Acc}_{\text{base}}},
\]
where $\text{Acc}_{\text{base}}$ denotes the task-specific chance baseline. We use the same normalization for the linguistic classification tasks in Section~\ref{sec:rq2}.

\paragraph{Zero-shot retrieval.}
To complement probe-based analyses, we evaluate zero-shot cross-modal retrieval on the same datasets using cosine similarity without training additional classifiers, reporting layer-wise Recall@K.
 
\paragraph{Relational retrieval.}

While cross-modal classification tests whether models map isolated words to their visual referents, it leaves open whether this alignment extends to compositional, attribute-level meaning. We introduce controlled relational retrieval tasks using synthetically generated scenes composed of fruit objects \cite{oltean2017fruits360} targeting three attributes: cardinality (e.g., ``3 apples''), spatial direction (e.g., ``apple right of banana''), and relative size (e.g., ``tiny orange''). This dataset consists of isolated fruit images against clean backgrounds, making it straightforward to compose controlled scenes by combining and manipulating individual objects to express target relations. These tasks pair natural images with rendered text-image descriptions of the target relation, enabling us to test whether cross-modal alignment extends beyond simple object-label correspondence to relational and attribute-level structure. We evaluate retrieval in both image-to-text and text-to-image directions and report average Recall@K.
 
\paragraph{Orthographic control.}

Cross-modal alignment could in principle reflect orthographic co-occurrence rather than semantic grounding. That is, a model might associate an image of a horse with ``horse'' simply because they co-occur in training, not because it understands the concept. We therefore curate 90 orthographic neighbor pairs corresponding to a subset of the 143 Caltech-256 classes, each pairing a class name with a semantically unrelated distractor at edit distance 1 (e.g., ``horse'' vs.\ ``house''). For each natural image, we report the cosine similarity gap between the correct and distractor rendered texts:

$$\Delta_{\cos} = \cos(E_{\text{text}}, E_{\text{img}}) - \cos(E_{\text{ortho}}, E_{\text{img}}),$$

where $E_{\text{img}}$, $E_{\text{text}}$, and $E_{\text{ortho}}$ are the embeddings of the image, correct text, and orthographic distractor, respectively. Positive values indicate that the model favors the semantically correct match over the orthographically similar one.

\subsection{Results and Discussion}
 
\begin{figure}[t]
\centering
\includegraphics[width=\linewidth]{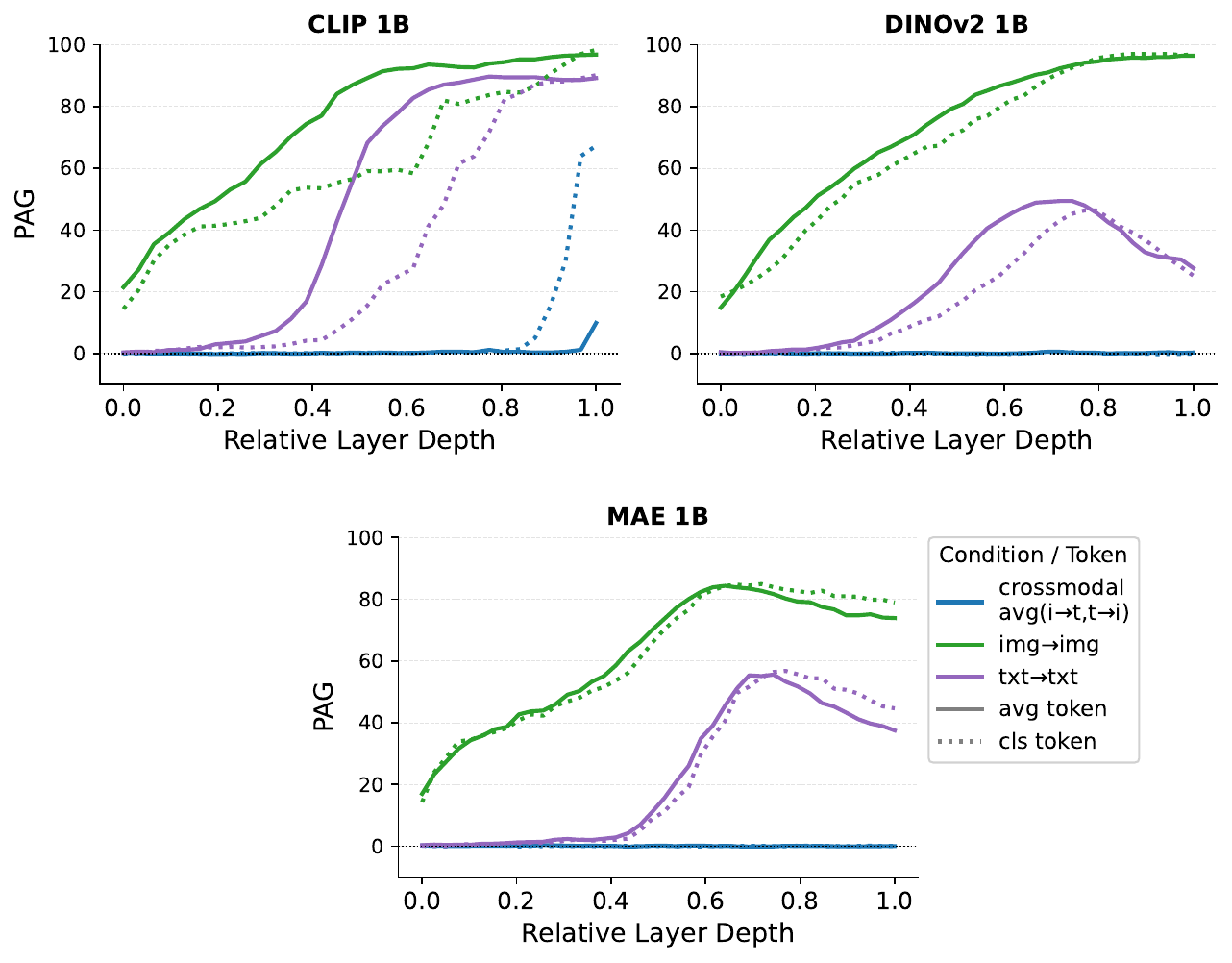}
\caption{Layer-wise PAG across 1B vision encoders for cross-modal classification. In CLIP, image recognition emerges first, followed by text recognition, and finally cross-modal transfer in the deepest layers. Cross-modal transfer remains near baseline in DINO and MAE.}
\label{fig:cross_uni}
\end{figure}

Across all models, image recognition (img-img) develops before text recognition (txt-txt), and cross-modal transfer, when present, emerges only after both, in the deepest layers (Figure~\ref{fig:cross_uni}). In CLIP, cross-modal transfer increases sharply in the final layers, whereas in DINO and MAE it remains near baseline throughout depth. In CLIP specifically, img-img scores rise steeply from depth 0.4, txt-txt scores follow from depth 0.6, and cross-modal transfer only begins to noticeably increase after depth 0.9. This suggests that shared cross-modal structure is not reducible to unimodal competence alone, but instead emerges only once image and text representations become sufficiently expressive. In DINO and MAE, both unimodal settings reach high PAG by depth 0.6--0.7, yet cross-modal transfer remains near baseline across all layers, indicating that strong unimodal representations alone are insufficient for robust cross-modal alignment. Results on ImageNet-1K (Appendix~\ref{appx:cross_imagenet}) show the same pattern. Supporting zero-shot retrieval experiments exhibit the same late-layer trend, with CLIP developing strong retrieval alignment while DINO and MAE (1B) remain near chance across depth (Appendix~\ref{appx:zeroshot}). Taken together, these results point to a consistent ordering: cross-modal alignment does not develop in parallel with unimodal recognition, but emerges only after both image and text representations have individually matured — suggesting it depends on the prior consolidation of unimodal competence rather than developing alongside it.

 \begin{figure}[t]
\centering
\includegraphics[width=\linewidth]{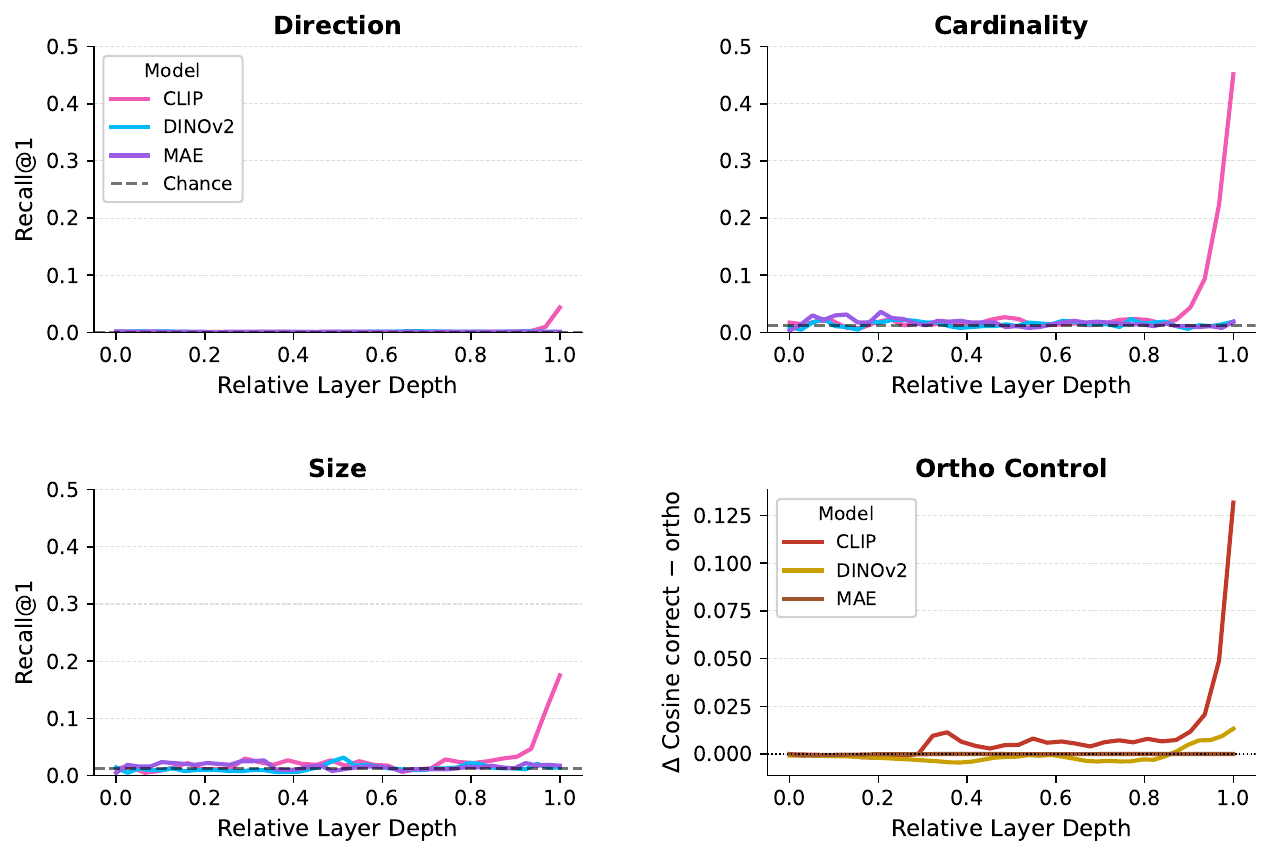}
\caption{Layer-wise Recall@1 for relational retrieval tasks. CLIP develops strong late-layer alignment for direction, cardinality, and size, while DINOv2 and MAE remain near chance across depth.}
\label{fig:relational_ortho}
\end{figure}
Relational retrieval experiments further show that this alignment extends beyond isolated object-label correspondence to attributes such as direction, size, and cardinality (Figure~\ref{fig:relational_ortho}, first 3 subplots). While absolute Recall@1 scores remain low across all tasks, CLIP consistently rises well above chance in the deepest layers, confirming that late-layer representations carry meaningful relational structure despite the overall difficulty of these tasks. Among the three attributes, cardinality shows the strongest alignment, followed by size, while spatial direction remains substantially weaker — suggesting that fine-grained relational correspondence is more difficult to recover than object-level semantic alignment. This is consistent with prior work showing that CLIP struggles broadly with spatial direction \citep{liu2023visual}, likely because directional prepositions are systematically underspecified in the web-scraped image-caption pairs used for contrastive pretraining.
 
Orthographic control experiments (Figure~\ref{fig:relational_ortho}, last subplot) further show that in early and middle layers, all three models assign similarly high cosine similarity to both the correct text and the orthographic distractor, indicating that representations at these depths are not yet semantically discriminative. In the deepest layers, however, CLIP shows a clear divergence: similarity to the correct text drops less sharply than similarity to the distractor, opening a growing gap that favors the semantically corresponding match. DINO and MAE show no such separation, with correct and distractor similarities tracking each other closely throughout depth. 

Across all three evaluations — cross-modal classification, relational retrieval, and orthographic control — CLIP consistently outperforms DINO and MAE, and in each case the advantage is concentrated in the deepest layers. This pattern is consistent with a role for contrastive language supervision: CLIP is explicitly trained to align image and text representations through contrastive language supervision, whereas DINO and MAE receive no such signal, suggesting that, at the 1B scale, robust cross-modal transfer, relational alignment, and semantic disambiguation beyond orthographic similarity depend strongly on language-aligned training signals; Section~\ref{sec:rq3} examines whether scale and data composition can compensate.
 
 \begin{figure}[t]
\centering
\includegraphics[width=\linewidth]{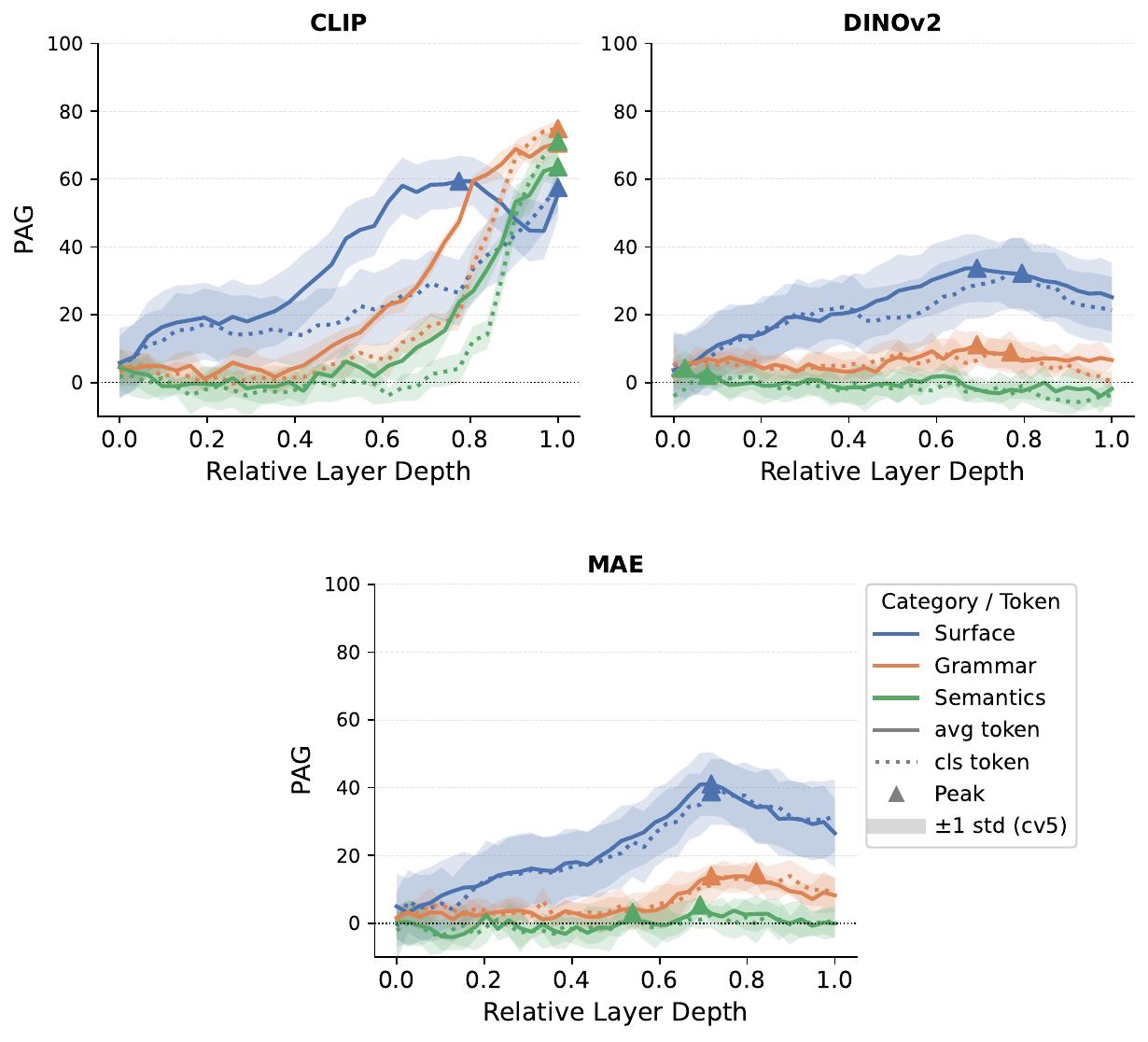}
\caption{Layer-wise PAG across 1B encoders for linguistic tasks. CLIP shows a clear surface-to-semantic progression; DINO and MAE show weak grammatical and near-absent semantic signals.}
\label{fig:hierarchy}
\end{figure}

\section{RQ2: Representational Structure}
\label{sec:rq2}

\subsection{Tasks}

 We examine which linguistic properties are recoverable from frozen representations at different network depths. Each example consists of a rendered word or phrase paired with a task-specific label. Given its representation at a particular layer, a probe predicts an orthographic, grammatical, or semantic property, as defined below.

\paragraph{Surface probes.}
We assess character identity via 26 binary classifiers (one per letter) and word length via length bin prediction. For \textit{character presence}, each classifier is trained on 100 positive examples (words containing the target letter) and 100 negative examples (words not containing it), balanced by selecting the most frequent words from SUBTLEX-US \citep{brysbaert2009moving}; each word is rendered as a single $224 \times 224$ image. For \textit{letter count}, we define four word-length bins ([3–5], [6–8], [9–11], [12–14] characters) with 300 words per bin (100 per individual length), again frequency-ordered from SUBTLEX-US, giving a 4-class problem with a chance level of 25\%. These tasks establish whether the encoder captures the most basic orthographic properties of rendered text, which are the building blocks of any deeper linguistic organization. For Letter Count, word length is directly recoverable from image width, introducing a visual confound that could inflate probe performance independently of any orthographic content. To discount this, we apply rectangle input controls, which are blank rectangles rendered in the same style and position as the original word, with width proportional to character count at the rendered font size and height equal to the text bounding box but containing no linguistic content (Figure~\ref{fig:input_control}, bottom row), and compute a confound PAG from their performance. The final adjusted score $\text{PAG}_{\text{adj}}$ subtracts this confound, ensuring that reported performance reflects orthographic encoding rather than incidental visual cues.

\paragraph{Grammatical probes.}
We use part-of-speech (POS) prediction as a proxy for grammatical organization beyond surface form. If representations encode not just what a word looks like but what syntactic role it plays, this would suggest that the encoder has internalized structure that goes beyond orthographic pattern matching. For \textit{isolated-word POS tagging}, we build a 4-class dataset (NOUN, VERB, ADJ, ADV) of up to 200 words per class drawn from five Universal Dependencies English treebanks \citep{nivre-etal-2020-universal} (EWT, GUM, LinES, ParTUT, PUD), retaining only unambiguously tagged lemma; each word is rendered as a single image. For \textit{contextual POS}, we target 4 classes (NOUN, VERB, ADJ, ADV) using words that appear under multiple POS tags across the same treebanks, collecting up to 500 instances per class; each image shows a 5-word context window with the target word circled, which is intended to reduce reliance on lexical priors.

\paragraph{Semantic probes.}
We assess whether representations capture higher-level conceptual structure through two tasks: \textit{supersense categorization}, which tests whether words are grouped by broad semantic category, and \textit{concreteness prediction}, which tests whether the encoder distinguishes between abstract and concrete concepts. For supersense, we define 10 WordNet-derived categories (animal, fruit, vegetable, furniture, vehicle, clothing, instrument, building, tool, tree) with up to 50 words per class, filtered by SUBTLEX-US frequency; each word is rendered as a single image. For concreteness, words are drawn from SUBTLEX-US  and labeled as abstract (concreteness rating $\leq 2.0$) or concrete ($\geq 3.0$) based on the concreteness ratings from \citet{brysbaert2014concreteness}, with the ambiguous mid-range dropped; 500 words per class are selected with word-length distributions matched across classes to remove length as a confound. Together, these probe whether the encoder has moved beyond surface and grammatical organization toward something closer to meaning. Rectangle input controls are applied to concreteness prediction, where word length is a known correlate of abstractness \citep{lewis2016length}.

\paragraph{Probe training and evaluation.}
For each task, an example consists of a frozen representation of a rendered word or phrase paired with its task-specific label. We evaluate independent L2-regularized logistic regression probes for each vision encoder, layer, and pooling method using stratified five-fold cross-validation with random shuffling. 
We standardize features using the mean and variance estimated from the training split. We select the regularization strength $C \in \{10^{-3}, 10^{-2}, 10^{-1}, 1, 10, 100\}$ by maximizing accuracy on a held-out validation subset comprising 10\% of the training split, then refit the classifier on the entire training split. We report mean test accuracy and standard deviation across the five folds and normalize accuracy using PAG (see \S~\ref{subsec:tasks}). For tasks with rectangle input controls, we report $\text{PAG}_{\text{adj}} = \text{PAG} - \text{PAG}_{\text{confound}}$, where $\text{PAG}_{\text{confound}}$ denotes probe performance on rectangle controls. 

To ensure probe performance reflects structured information rather than superficial cues, we apply label controls (shuffled labels, fixed inputs) to all tasks \citep{hewitt2019designinginterpretingprobescontrol}; as control performance remained consistently near chance level, we exclude them from further analysis. To reduce task-specific variance in reporting, we aggregate probing performance within each linguistic level into a single composite PAG. Full dataset construction details are in Appendix~\ref{appx:dataset}.

\begin{figure*}[t]
\includegraphics[width=\linewidth]{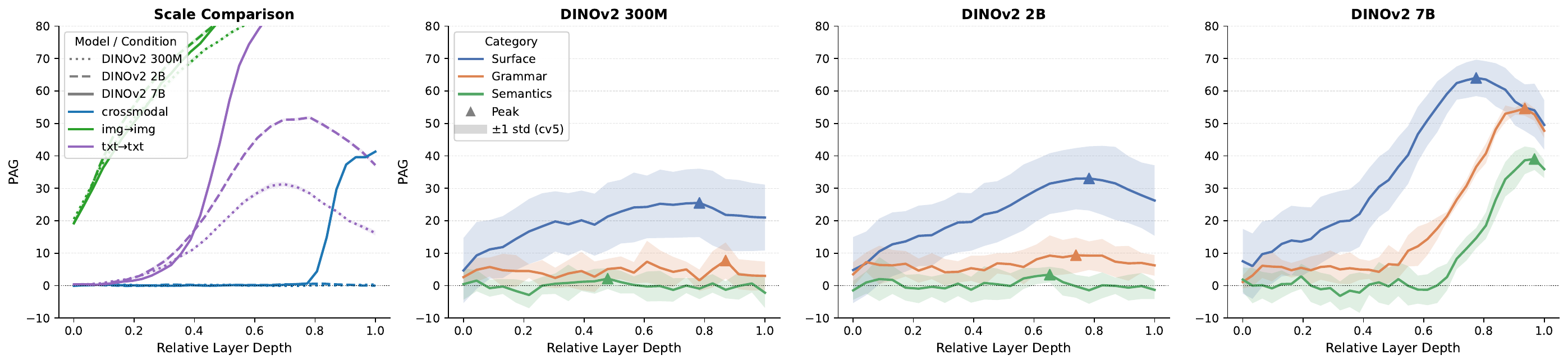}
\caption{\textit{Left (1)}: Cross-modal and unimodal classification across scales. Scale substantially improves cross-modal alignment. \textit{Right (2-4)}: Layer-wise PAG across DINO scales (300M, 2B, 7B) for linguistic probing. Larger models, with the 7B model also trained on more data, exhibit stronger and more hierarchically organized representations.}
\label{fig:dino_scale}
\end{figure*}

\begin{figure*}[t]
\includegraphics[width=\linewidth]{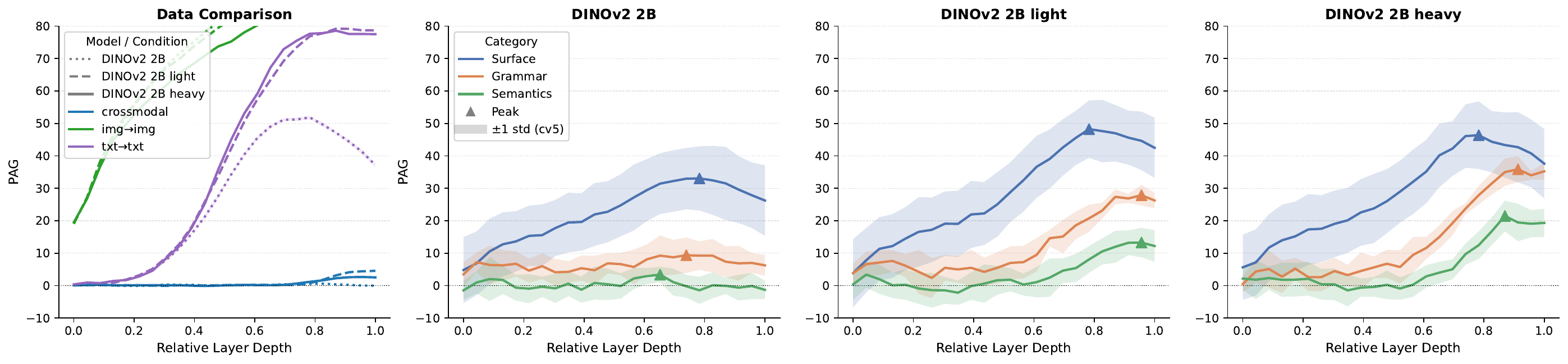}
\caption{\textit{Left (1)}: Cross-modal and unimodal classification across data compositions: Normal, light (50.3\% text only), heavy (1.3\% OCR only). Data composition modestly improves cross-modal alignment. \textit{Right (2-4)}: Layer-wise PAG across DINO pretrained on different data compositions. OCR-focused data strengthens and hierarchically organizes representations.}
\label{fig:dino_data}
\end{figure*}

\subsection{Results and Discussion}

Figure~\ref{fig:hierarchy} shows layer-wise PAG for surface, grammatical, and semantic probing tasks across 1B models. CLIP develops a clear surface-to-semantic hierarchy across layers: surface information emerges early, grammatical regularities peak at intermediate depth, and semantic organization consolidates only in the deepest layers (depth $\approx$0.9). Similar trends are observed in the contextual POS setting (Appendix~\ref{appx:phrase_pos}), confirming that these grammatical signals are not limited to isolated lexical priors. In contrast, DINO and MAE exhibit strong surface signals but substantially weaker grammatical and semantic structure.
 
Importantly, the emergence of semantic probe performance in CLIP coincides with the onset of strong cross-modal transfer and retrieval alignment from RQ1. Orthographic control experiments further show that late-layer image-text similarity increasingly favors correct semantic matches over orthographically similar distractors, indicating that late-layer alignment in CLIP corresponds to increasingly structured semantic organization rather than purely orthographic matching. These findings corroborate \citet{materzynska2022disentanglingvisualwrittenconcepts} while extending their results by localizing strong cross-modal alignment to the deepest layers and relating it to a progressively organized representational hierarchy. They also partially support the concerns of \citet{wang2025textural}: orthographic similarity contributes substantially in earlier layers, but later CLIP representations increasingly reflect semantic correspondence beyond surface form.
 
The contrast with DINO and MAE is informative: both models develop strong surface representations from visual form alone, but, at this scale, the absence of language supervision limits the emergence of the deeper grammatical and semantic structures that support cross-modal alignment in CLIP.

\section{RQ3: The Role of Scale and Data Composition}
\label{sec:rq3}

\subsection{Tasks}
 
Having established that cross-modal alignment and its underlying representational hierarchy depend strongly on training objective, we examine whether scale and data composition can compensate for the absence of language supervision. We study a family of self-supervised DINO variants that differ either in model size or training data composition. To examine how model scale and training-data composition affect alignment and representational structure, we apply the tasks and evaluation protocols introduced in Sections~\ref{sec:rq1} and~\ref{sec:rq2} to the DINO variants described below.
 
\paragraph{Scale variants.}
We compare DINO models with 300M, 2B, and 7B parameters to examine how scaling is associated with cross-modal transfer and representational organization. The 300M and 2B models are trained on the same 2B-image dataset, whereas the 7B model is trained on a larger 8B-image dataset. Thus, comparisons between the 300M and 2B models isolate the effect of model capacity, while differences involving the 7B model reflect the combined effects of increased capacity and training-data scale.
 
\paragraph{Data composition variants.}
We examine DINO-2B models trained on subsets of the MC dataset \cite{xu2024demystifying} enriched for text-heavy content. The ``light'' variant retains 50.3\% of images in total through mild OCR-based filtering, while the ``heavy'' variant retains only 1.3\%, primarily documents, charts, and tables.
 
\subsection{Results and Discussion}

\paragraph{Effect of model scale.}
 Figure~\ref{fig:dino_scale} shows that scale improves both cross-modal transfer and representational organization, with the largest gains at 7B. While image-to-image alignment improves steadily across depth for all models, text-to-text recognition is less stable in smaller models, showing gains in intermediate layers but degrading in deeper layers for the 300M and 2B variants. In contrast, the 7B model maintains strong text representations in later layers and first exhibits clear cross-modal transfer in deep layers among self-supervised models.
 
Surface-level information is learned early across all scales, with larger models consistently performing better. Increasing scale also produces a clearer separation between surface, grammatical, and semantic representations across depth, with semantic probe performance strengthening substantially in the deepest layers. This hierarchical organization remains weak in smaller models, where semantic probe performance stays near baseline. Zero-shot retrieval, relational retrieval, and orthographic control experiments exhibit the same trend (Appendix~\ref{appx:zeroshot}, \ref{appx:relational_ortho}), with DINOv2 7B showing substantially stronger late-layer alignment and semantic separation than smaller self-supervised models.

\paragraph{Effect of data composition.}
Prior work by \citet{fan2025scalinglanguagefreevisualrepresentation} showed that scale and text-rich data can partially offset the OCR advantage provided by language supervision. Figure~\ref{fig:dino_data} shows that this effect extends beyond surface-level recognition into cross-modal transfer and representational organization. Filtering toward text-rich data stabilizes text-to-text representations in deeper layers. The baseline model exhibits a noticeable degradation in late-layer text recognition that text-rich filtering corrects. Filtering also produces modest improvements in late-layer cross-modal transfer and semantic probe performance. However, these gains remain substantially weaker than those observed in DINOv2 7B and CLIP. Supporting retrieval and orthographic control experiments show the same qualitative trend (Appendix~\ref{appx:zeroshot}, \ref{appx:relational_ortho}), suggesting that data composition alone is insufficient to fully recover the stronger cross-modal alignment observed in contrastively trained or large-scale models.
 
Overall, both increased model capacity and training-data scale are associated with stronger hierarchical linguistic organization and cross-modal alignment in self-supervised vision encoders. The controlled 300M-to-2B comparison shows that capacity alone mainly improves surface-level representations, with little effect on cross-modal transfer or semantic organization, while the stronger performance of DINOv2 7B reflects the combined effects of increased capacity and a larger training dataset. Optimized data composition improves text recognition and reduces modality gaps, but, in our experiments, does not match the transfer or semantic organization observed in the 7B model.

\section{Conclusion}

In this work, we studied how vision encoders process and organize textual information across network depth, using cross-modal alignment between rendered text and visual concepts as a controlled lens into broader reading comprehension capabilities. Through controlled cross-modal transfer experiments and supporting layer-wise probing analyses, we showed that strong cross-modal alignment between rendered text and natural images emerges only in the deepest layers, after unimodal object and text recognition have already developed. Zero-shot retrieval and relational retrieval experiments corroborate this picture, showing that late-layer CLIP representations support not only object-label correspondence but also attribute-level structure such as cardinality and size, while spatial direction remains consistently the most difficult relational attribute to recover. Our analyses further revealed that this transition coincides with a progressively organized hierarchy of representations spanning surface-level orthographic information, grammatical regularities, and semantic organization. Our findings suggest that strong cross-modal semantic alignment in vision encoders emerges jointly from sufficient model capacity and exposure to language-aligned visual signals. Among the models studied, increasing model size and training-data scale yield larger gains in cross-modal transfer and hierarchical linguistic organization than enriching training data with text does. Text-rich data stabilizes late-layer text recognition and modestly improves semantic alignment, but does not match the stronger transfer and semantic organization observed in the 7B model.

\section*{Limitations}

Our work studies language-relevant representations in vision encoders primarily through cross-modal semantic alignment between rendered text and visual concepts. While this provides a measurable operational proxy, reading comprehension is substantially broader and more complex than the behaviors evaluated here, involving compositional reasoning, discourse understanding, world knowledge integration, long-range dependency processing, and pragmatic interpretation. As a result, our findings should be interpreted as evidence of language-relevant representational organization and cross-modal alignment rather than full human-like reading comprehension.

Our analyses also rely heavily on probing methods, which measure information accessible within intermediate representations rather than the causal mechanisms used during inference. Although we complement probing with cross-modal transfer, retrieval, and orthographic control experiments, these methods cannot fully determine how models internally compute or use linguistic representations. Moreover, many of our probing tasks operate primarily on isolated words. While this enables controlled analysis of orthographic, grammatical, and semantic structure, natural language understanding depends heavily on broader contextual and compositional information. We partially address this through contextual POS experiments, but more complex sentence-level and discourse-level evaluations remain outside the scope of this work.

Finally, our study focuses exclusively on standalone vision encoders rather than full large vision-language models (LVLMs) that integrate these encoders within larger multimodal systems. Since downstream language models may substantially alter or amplify the language-relevant representations present in the visual backbone, it remains unclear how these findings transfer to instruction-tuned or generative multimodal systems, which we leave for future work.

\section*{Acknowledgments}
This project has received funding from the European Research Council (ERC) under the European Union’s Horizon 2020 research and innovation programme (grant agreement No 101122047).

\section*{Disclaimers}
\paragraph{Use of existing artifacts.}
For this study, we rely on preexisting models and datasets; their use is consistent with their respective licenses.
\paragraph{Use of AI tools.}
The authors used AI tools to support code development and debugging. All experimental design, analyses, and scientific conclusions are solely the work of the authors, who take full responsibility for the contents of this paper.

\bibliography{ref}

\appendix
\section{Dataset Construction Details}
\label{appx:dataset}

\subsection{Character Presence}

Word stimuli are drawn from the SUBTLEX-US corpus, a subtitle-derived lexical database commonly used in psycholinguistic research. Candidate words are sorted by \texttt{Lg10CD}, a log-transformed contextual diversity measure reflecting the number of subtitle contexts in which a word appears. We remove words beginning with uppercase characters, words with length $\leq 2$, and words containing non-alphabetic characters such as apostrophes or hyphens. All remaining words are lowercased and deduplicated.

For each character $c \in \{a,\dots,z\}$, we construct a balanced binary dataset consisting of 100 high-frequency words containing $c$ and 100 high-frequency words not containing $c$. Words are selected deterministically through a greedy pass over the frequency-sorted corpus without random sampling. The resulting dataset contains 200 samples per character and 5,200 rendered images in total.

\subsection{Letter Count}

Word stimuli are drawn from the same filtered SUBTLEX-US vocabulary used for Character Presence. Words are grouped into four length bins: 3--5, 6--8, 9--11, and 12--14 characters. To ensure balanced class distributions, we select the top 100 most frequent words for each exact word length using a deterministic greedy pass over the frequency-sorted corpus. Each bin therefore contains 300 words (3 lengths $\times$ 100 words), resulting in 1,200 total word samples.

For every word, we generate two aligned image variants: a rendered text image and a rectangle control image. The rectangle image preserves the same rendering style, position, color, and approximate visual extent as the original word image, but replaces the rendered text with a filled rectangle whose width is proportional to the number of characters in the word. This control isolates visual span information while removing orthographic content.

\subsection{POS Tagging}

Word stimuli are constructed from five English Universal Dependencies (UD) treebanks: EWT, GUM, LinES, ParTUT, and PUD. All train, development, and test splits are merged into a single corpus spanning diverse domains including web text, news, fiction, subtitles, parliamentary proceedings, and Wikipedia. We parse all CoNLL-U files and remove comment lines, multiword tokens, empty nodes, punctuation, and non-alphabetic forms. All remaining word forms and lemmas are lowercased.

For each surviving token, we collect $(\text{form}, \text{lemma}, \text{UPOS})$ tuples and aggregate POS statistics at the lemma level across the merged corpus. To construct an unambiguous POS dataset, each lemma must occur at least 10 times overall and exhibit a dominant POS purity of at least 98\%, computed as the fraction of occurrences assigned to its most frequent POS tag. This filtering removes words with substantial POS ambiguity across contexts.

We retain four content-word POS categories: NOUN, VERB, ADJ, and ADV. Function-word categories such as determiners and adpositions are excluded because they are often trivially identifiable from orthographic form alone. For each retained POS class, we select the top 200 lemmas by corpus frequency, resulting in 800 total word samples.

\paragraph{Contextual POS Tagging.}

To complement isolated-word POS tagging, we additionally construct a contextual POS dataset using the same five Universal Dependencies treebanks. Unlike the primary POS dataset, this variant focuses exclusively on contextually ambiguous lemmas that occur under multiple POS categories in the corpus. A lemma is retained if it appears under at least two target POS classes, each occurring at least three times, with a minimum total frequency of 10 occurrences overall.

For each ambiguous token occurrence, we extract a short phrase centered on the target word using a five-word context window consisting of the two nearest content words on either side. Punctuation between anchor tokens is preserved to maintain natural sentence structure. To reduce duplication, examples are deduplicated by phrase content, lemma, and POS assignment.

We retain the same four POS categories used in the main experiments (NOUN, VERB, ADJ, ADV) and sample up to 500 examples per class, yielding at most 2,000 contextual phrase examples overall. Within each phrase image, the target word is visually highlighted with an ellipse to indicate the prediction target while preserving surrounding contextual information.

\subsection{Supersense Classification}

The supersense dataset consists of manually curated word lists spanning 10 superordinate semantic categories: animal, fruit, vegetable, furniture, vehicle, clothing, instrument, building, tool, and tree. Each category contains 50 concrete noun concepts selected to represent a broad range of category members, including both common and less frequent terms. Category definitions are loosely inspired by high-level WordNet semantic groupings, and the 50 words are filtered by SUBTLEX-US frequencies.

\subsection{Concreteness}

Concreteness labels are derived from the concreteness norms of \citet{brysbaert2014concreteness}, which provide crowd-sourced ratings for approximately 40,000 English lemmas on a 1--5 scale ranging from highly abstract to highly concrete concepts. We retain only alphabetic single-word entries, remove multi-word expressions and hyphenated forms, lowercase all words, and deduplicate the resulting vocabulary.

Words with concreteness ratings $\leq 2.0$ are assigned to the abstract class, while words with ratings $\geq 3.0$ are assigned to the concrete class. Intermediate words with ratings between 2.0 and 3.0 are excluded to avoid semantically ambiguous boundary cases.

Because concreteness is correlated with word length, we additionally perform length-matched balancing between the two classes. For each word-length bucket, we compute the maximum number of matched abstract-concrete pairs available and allocate samples proportionally across lengths. Within each $(\text{length}, \text{class})$ group, words are selected by descending SUBTLEX-US \texttt{Lg10CD} frequency. This procedure ensures that abstract and concrete classes exhibit identical word-length distributions, reducing the possibility of exploiting visual word extent as a shortcut.

The final dataset contains 500 abstract and 500 concrete words, for a total of 1,000 samples.

\subsection{Cross-Modal Classification}

Cross-modal classification evaluates whether representations learned from one modality generalize to the other. We construct paired image-text datasets using Caltech-256 and ImageNet-1K. For Caltech-256, we retain 143 visually well-defined classes after removing ambiguous or underspecified categories. For ImageNet-1K, all 1,000 classes are used. Natural images are paired with rendered text images corresponding to their class labels.

Two transfer directions are evaluated independently: image-to-text (img2txt), where classifiers are trained on image representations and evaluated on text representations, and text-to-image (txt2img), where training and evaluation are reversed. Unlike standard train-test splits, the generalization being evaluated occurs entirely across modalities rather than across samples. For Caltech-256, image-to-text training uses approximately 7,150 image samples with 143 text-image test samples, while ImageNet-1K uses 50,000 image samples with 1,000 text-image test samples.

\subsection{Orthographic Control}

The orthographic control dataset is constructed to test whether cross-modal alignment between natural images and rendered text can be explained purely through orthographic similarity. We manually curate 90 orthographic neighbor pairs for 90 of the 143 Caltech-256 classes used in the main experiments, namely those for which a semantically unrelated edit-distance-1 neighbor exists. Each pair consists of a class name and an orthographically similar distractor word with edit distance 1, generated through either single-character substitution or deletion (e.g., \textit{horse} $\rightarrow$ \textit{house}, \textit{goose} $\rightarrow$ \textit{noose}). Distractor words are selected to be semantically unrelated to the original class while remaining visually similar at the character level.

For each class, we construct three aligned modalities: natural images from Caltech-256, rendered text images of the correct class label, and rendered text images of the orthographic distractor. Orthographic control images are generated with the same canvas dimensions as their corresponding class-label renders to ensure that differences arise only from character content rather than image size or layout.

\subsection{Relational Retrieval}

Relational retrieval datasets are constructed using object instances from Fruits-360 \cite{oltean2017fruits360}. We select 20 fruit categories and generate synthetic image-text pairs targeting three relational attributes: cardinality, relative size, and spatial direction. Each dataset defines compound labels combining object identity with the target relational attribute.

For the cardinality dataset, each sample contains between one and four instances of a fruit object arranged on a fixed canvas. Compound labels therefore take the form \textit{apple\_2}, \textit{banana\_4}, etc. Object layouts vary according to count, while individual samples use different rotation frames from Fruits-360 together with small positional jitter to increase visual diversity. Corresponding text captions follow forms such as ``2 apples'' or ``1 apple''. The resulting dataset contains 80 compound labels with 400 image samples and 400 text samples overall.

For the size dataset, individual fruit instances are resized to one of four relative scales: tiny, small, big, or huge. Compound labels therefore correspond to object-size pairs such as \textit{apple\_tiny} or \textit{orange\_huge}. Text captions follow the corresponding linguistic form (e.g., ``tiny apple''). As in the cardinality dataset, multiple rotation frames and positional jitter are used for variation. The final dataset contains 80 compound labels with 400 image samples and 400 text samples.

For the direction dataset, we generate ordered object pairs across all distinct fruit combinations, producing compound labels such as \textit{apple\_right\_banana}. Four spatial relations are used: top, right, bottom, and left. Objects are positioned symmetrically around the image center with independent positional jitter applied to each object. Corresponding text captions follow forms such as ``apple right of banana''. This dataset contains 1,520 relational triples with 7,600 image samples and 7,600 text samples overall.


\section{Models}
\label{appx:models}
\begin{table}[H]
  \centering
  \small
  \setlength{\tabcolsep}{4pt}
  \begin{tabular}{ll}
  \toprule
  \textbf{Identifier} & \textbf{Objective} \\
  \midrule
  \multicolumn{2}{l}{\textit{300M}} \\
  \quad\texttt{facebook/webssl-dino300m-full2b-224}        & DINO \\
  \midrule
  \multicolumn{2}{l}{\textit{1B}} \\
  \quad\texttt{facebook/webssl-dino1b-full2b-224}     & DINO \\
  \quad\texttt{facebook/webssl-mae1b-full2b-224}      & MAE  \\
  \quad\texttt{facebook/metaclip-h14-fullcc2.5b}       & CLIP \\
  \midrule
  \multicolumn{2}{l}{\textit{2B}} \\
  \quad\texttt{facebook/webssl-dino2b-light2b-224}  & DINO \\
  \quad\texttt{facebook/webssl-dino2b-heavy2b-224}   & DINO \\
  \quad\texttt{facebook/webssl-dino2b-full2b-224}  & DINO \\
  \midrule
  \multicolumn{2}{l}{\textit{7B}} \\
  \quad\texttt{facebook/webssl-dino7b-full8b-224 }     & DINO \\
  \bottomrule
  \end{tabular}
  \caption{Vision encoders used in this work. All models
  are available at \texttt{huggingface.co}.}
  \label{tab:models}
  \end{table}

\section{Further Results}
\subsection{Cross-modal Classification for ImageNet1k}
\label{appx:cross_imagenet}          

\begin{figure}[H]  
\centering      
\includegraphics[width=\linewidth]{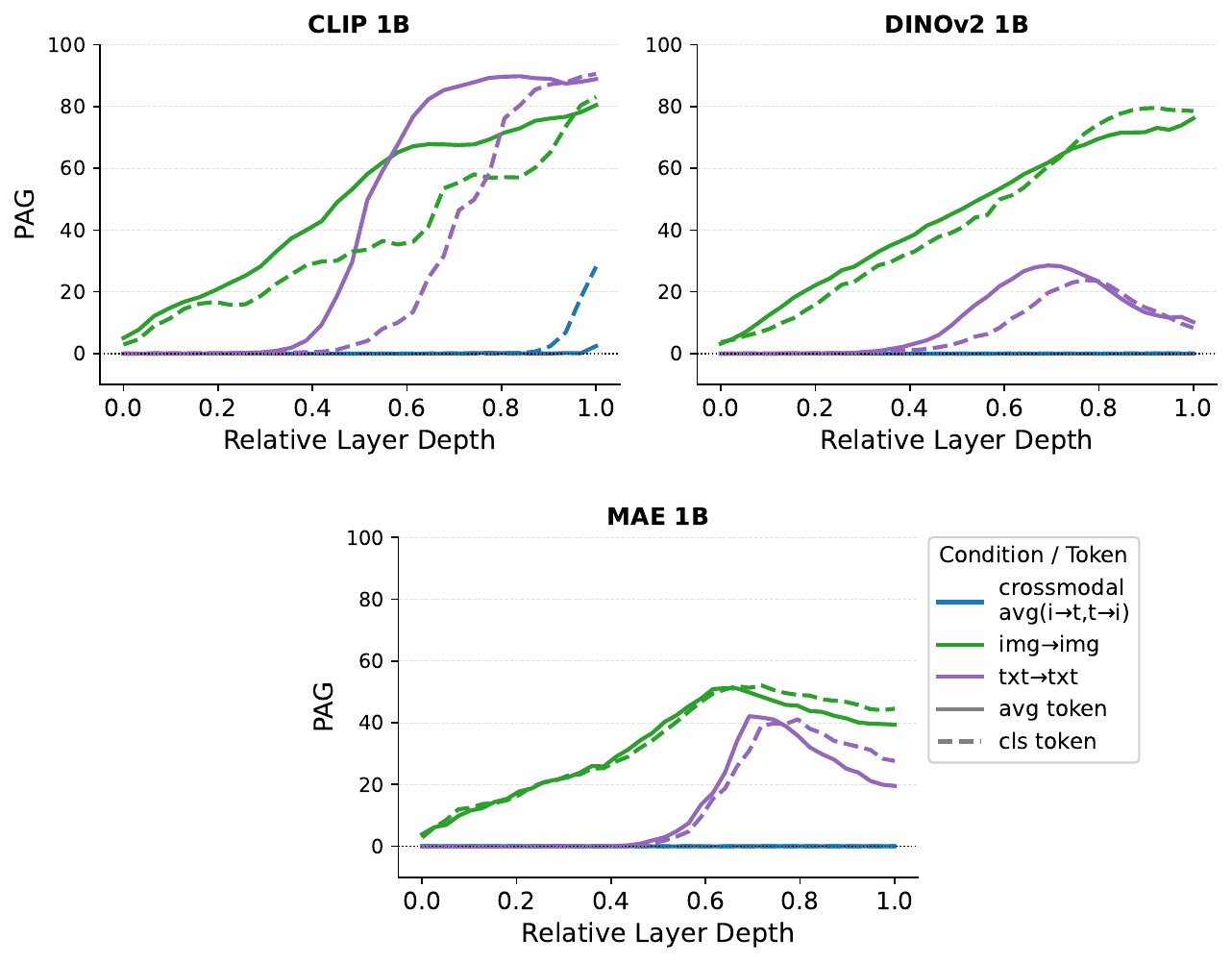}
\caption{Layer-wise PAG across 1B vision encoders for cross-modal classification on ImageNet1k. Similar to results on Caltech256, in CLIP, image recognition develops first, followed by text recognition, and finally cross-modal alignment, each peaking in deeper layers. Cross-modal transfer is absent in both DINO and MAE.}
\label{fig:cross_imagenet}          
\end{figure}  

\subsection{Zero-shot Retrieval}
\label{appx:zeroshot}

\begin{figure}[H]  
\centering      
\includegraphics[width=\linewidth]{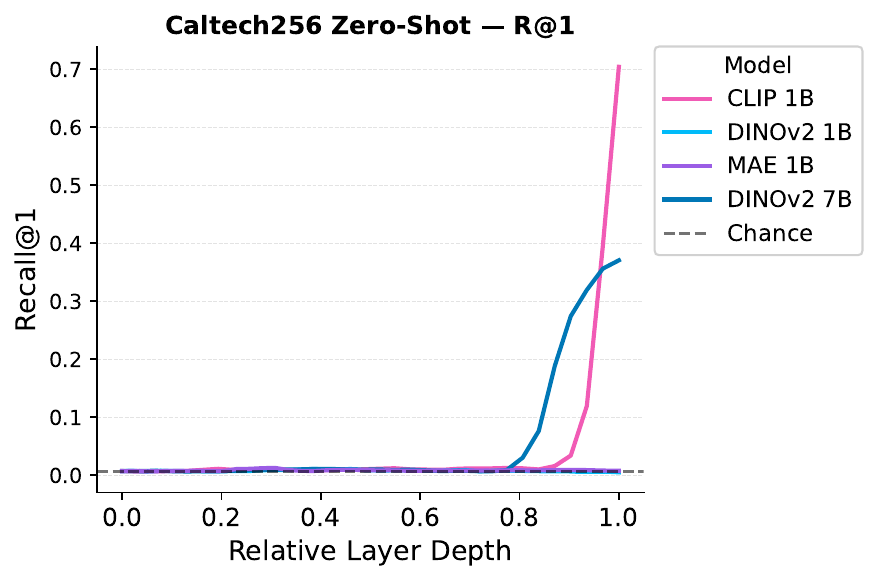}
\caption{Layer-wise zero-shot cross-modal retrieval on Caltech-256 measured with Recall@1. CLIP and DINOv2 7B exhibit sharp late-layer increases in retrieval alignment between rendered text and natural images, while DINOv2 1B and MAE remain near chance across depth.}
\label{fig:cross_imagenet}          
\end{figure}  

\begin{figure}[H]  
\centering      
\includegraphics[width=\linewidth]{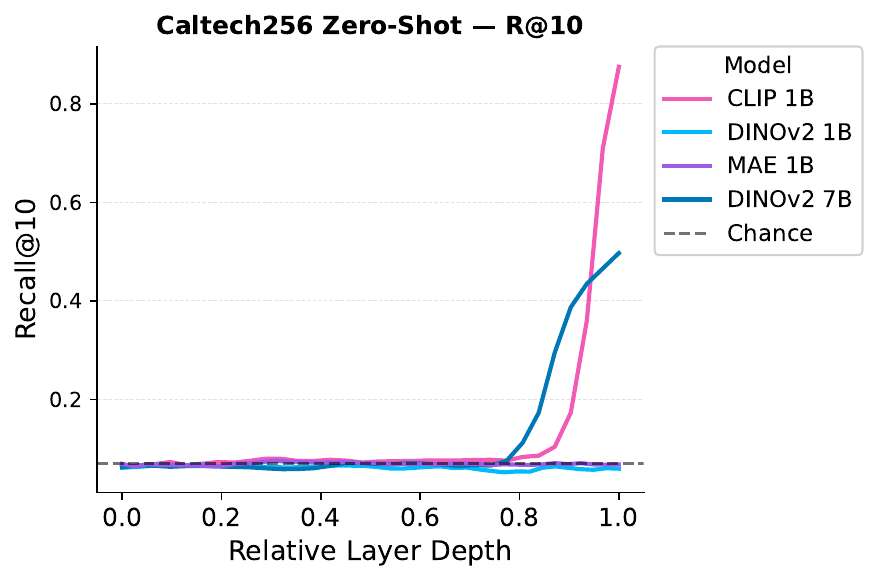}
\caption{Layer-wise zero-shot cross-modal retrieval on Caltech-256 measured with Recall@10. Similar to Recall@1, retrieval performance rises sharply only in the deepest layers of CLIP and DINOv2 7B, mirroring the late emergence of cross-modal transfer observed in probe-based experiments.}
\label{fig:cross_imagenet}          
\end{figure}  

\subsection{Relational Retrieval and Orthographic Control}
\label{appx:relational_ortho}
\begin{figure}[H]  
\centering      
\includegraphics[width=\linewidth]{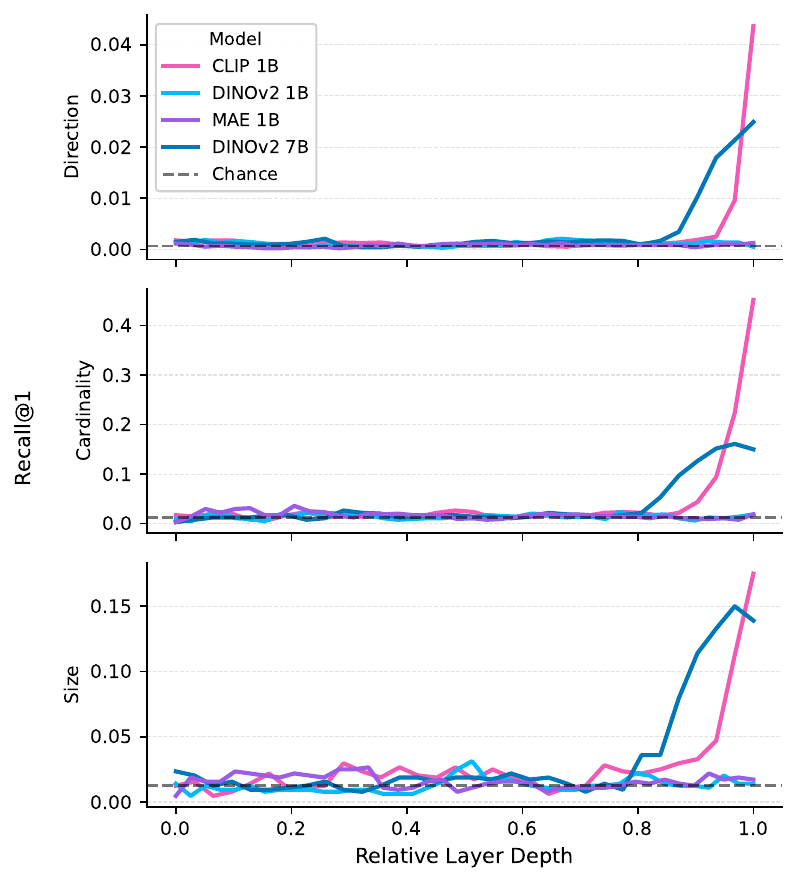}
\caption{Layer-wise zero-shot relational retrieval measured with Recall@1. CLIP develops strong late-layer alignment for direction, cardinality, and size, while DINOv2 7B exhibits weaker but measurable relational alignment in deeper layers. Smaller self-supervised models remain near chance across depth, with spatial direction consistently emerging as the most difficult relational attribute.}
\label{fig:cross_imagenet}          
\end{figure} 

\subsection{Contextual POS}
\label{appx:phrase_pos}
\begin{figure}[H]  
\centering      
\includegraphics[width=\linewidth]{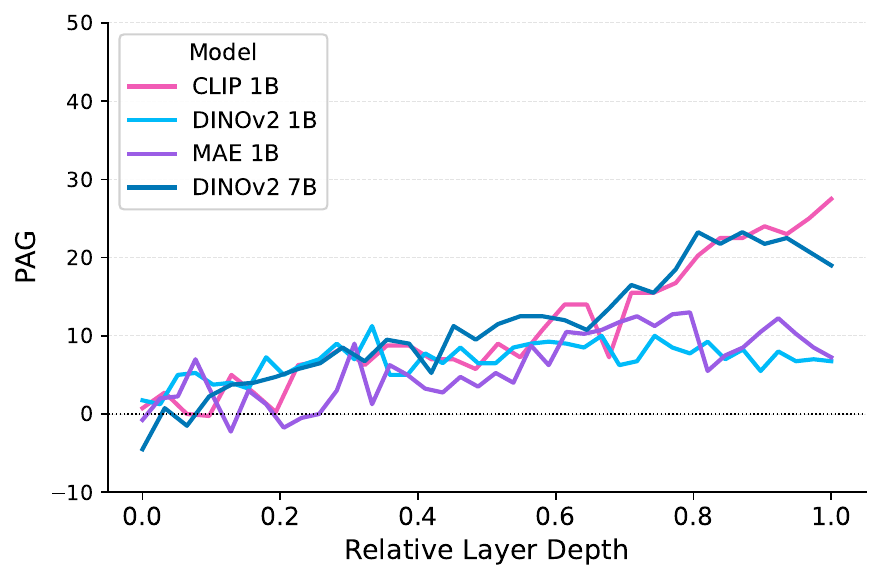}
\caption{Layer-wise PAG for contextual POS tagging across 1B vision encoders and DINOv2 7B. Similar to isolated-word POS probing, CLIP and DINOv2 7B exhibit progressively stronger grammatical signals in deeper layers, while DINOv2 1B and MAE remain substantially weaker.}       
\end{figure}

\end{document}